\documentclass[conference,compsoc]{IEEEtran}
\ifCLASSOPTIONcompsoc
  \usepackage[nocompress]{cite}
\else
  \usepackage{cite}
\fi
\ifCLASSINFOpdf
  \usepackage[pdftex]{graphicx}
\else
  \usepackage[dvips]{graphicx}
\fi
\usepackage{amsmath}
\ifCLASSOPTIONcompsoc
  \usepackage[caption=false,font=footnotesize,labelfont=sf,textfont=sf]{subfig}
\else
  \usepackage[caption=false,font=footnotesize]{subfig}
\fi
\usepackage{upquote}

\begin{document}
%
\title{3D Scanning System for Automatic High-Resolution Plant Phenotyping}

\author{
\IEEEauthorblockN{Authors to be added after review}
\IEEEauthorblockA{Affiliation\\
Address\\
Email: email}
}

\author{
\IEEEauthorblockN{Chuong V. Nguyen}
\IEEEauthorblockA{ARC Centre of Excellence for Robotic Vision\\ Research School of Engineering, Australian National University\\
Canberra ACT 2601, Australia\\
Email: Chuong.Nguyen@anu.edu.au}
\and
\IEEEauthorblockN{Jurgen Fripp}
\IEEEauthorblockA{CSIRO Health and Biosecurity\\
Australian eHealth Research Centre\\
Herston QLD 4029, Australia\\
Email: Jurgen.Fripp@csiro.au}
\and
\IEEEauthorblockN{David R. Lovell}
\IEEEauthorblockA{Electrical Eng. \& Computer Science\\
Queensland University of Technology\\
Brisbane QLD 4001, Australia\\
Email: David.Lovell@qut.edu.au}
\and
\IEEEauthorblockN{Robert Furbank}
\IEEEauthorblockA{CoE for Translational Photosynthesis\\ Australian National University \\ 
Canberra ACT 2601, Australia\\
Email: Robert.Furbank@anu.edu.au}
\and
\IEEEauthorblockN{Peter Kuffner, Helen Daily, Xavier Sirault}
\IEEEauthorblockA{CSIRO Agriculture and Food\\
High Resolution Plant Phenomics Centre\\
Canberra ACT 2601, Australia\\
Email: Xavier.Sirault@csiro.au}
}

\maketitle
\begin{abstract}
Thin leaves, fine stems, self-occlusion, non-rigid and slowly changing structures make plants difficult for three-dimensional (3D) scanning and reconstruction -- two critical steps in automated visual phenotyping. Many current solutions such as laser scanning, structured light, and multiview stereo can struggle to acquire usable 3D models because of limitations in scanning resolution and calibration accuracy. 
In response, we have developed a fast, low-cost, 3D scanning platform to image plants on a rotating stage with two tilting DSLR cameras centred on the plant. This uses new methods of camera calibration and background removal to achieve high-accuracy 3D reconstruction. We assessed the system's accuracy using a 3D visual hull reconstruction algorithm applied on 2 plastic models of dicotyledonous plants, 2 sorghum plants and 2 wheat plants across different sets of tilt angles. Scan times ranged from 3 minutes (to capture 72 images using 2 tilt angles), to 30 minutes (to capture 360 images using 10 tilt angles). The leaf lengths, widths, areas and perimeters of the plastic models were measured manually and compared to measurements from the scanning system: results were within 3-4\% of each other. The 3D reconstructions obtained with the scanning system show excellent geometric agreement with all six plant specimens, even plants with thin leaves and fine stems.
\end{abstract}


%
\IEEEpeerreviewmaketitle

\section{Introduction}
The structures of different plant species pose a range of challenges for 3D scanning and reconstruction. Various solutions to the issue of digitally imaging plants have been reported.  For plants with larger leaves and simple structures (such as maize, sorghum, cereal seedlings) it is possible to capture a small number of digital images from various viewing angles (typically 3), analyse these in 2D, then develop a relationship between these 2D poses and the (destructively measured) leaf area and biomass of the species \cite{Rajendran2009}. However, commercial systems using this approach are relatively expensive, with closed and proprietary analysis software; these have generally been deployed in large phenomics centres with conveyor based systems \cite{Danforth2016}. The 2D approach has difficulty in resolving concavities, leaf overlap and other occlusions; many of the powerful image analysis tools which can be applied to 3D meshes (e.g., volumetric based shape recognition and organ tracking) are more difficult in 2D \cite{Paproki2012}.  

Laser scanning, e.g., light detection and ranging (LIDAR), has been applied to plant digitisation but reconstructing a mesh from a pointcloud sufficiently dense to capture thin narrow leaves is computationally intensive. While this approach has been applied successfully to forestry \cite{Jupp2009, Yang2013} and to statistical analysis of canopies, it is not well suited to extracting single plant attributes \cite{Jupp2009, Deery2014}. Full waveform LIDAR is extremely expensive; simpler machine vision LIDAR systems of sufficient resolution can cost tens of thousands of dollars. Structured light approaches using affordable sensors such as the Kinect gaming sensor or camera-video projector setups do not offer the resolution or spatial repeatability to cope with complex plant structures \cite{Nguyen2012, Li2013}.  

Recently, approaches using multiple images from a larger number of viewing angles have yielded promising results \cite{Paproki2012, Sirault2013, Tabb2013, Lou2014, Pound2014, Nguyen2015structured}, either by using a silhouette-based 3D mesh reconstruction method or patch based stereo reconstruction transforming the images to point clouds. Silhouette-based reconstruction is prone to errors in camera calibration and silhouette extraction. Image acquisition for 3D reconstruction remains largely manual \cite{Duan2016}, limiting the speed and accuracy of 3D reconstruction.

Existing background removal techniques for silhouette extraction \cite{Lo2006, Campbell2011} are not reliable for complex plant structures. For current patch based methods,  reconstruction quality is usually poor due to both weak-textures (patterns on leaves) and thin structures. In both cases, high accuracy 3D reconstruction requires a very rigid imaging station and the engineering required for sensor integration is costly. Recent work has included semi-automated approaches to modelling the plants \cite{Quan2006, Yin2016} increasing completeness and visual quality at the expense of throughput. 

To address the shortcomings of existing scanning systems, we describe ``PlantScan Lite'', an affordable automated 3D imaging platform system to accurately digitise complex plant structures. This can be readily built for less than AU\$8000 of components (including DSLR cameras). The system has a number of novel elements including the use of high-resolution digital single-lens reflex (DSLR) cameras synchronised with a turntable; a high-accuracy, easy-to-setup camera calibration; and an accurate background removal method for 3D reconstruction.

\section{Methods}
\begin{figure}[!t]
\centering
\includegraphics[width=3.5in]{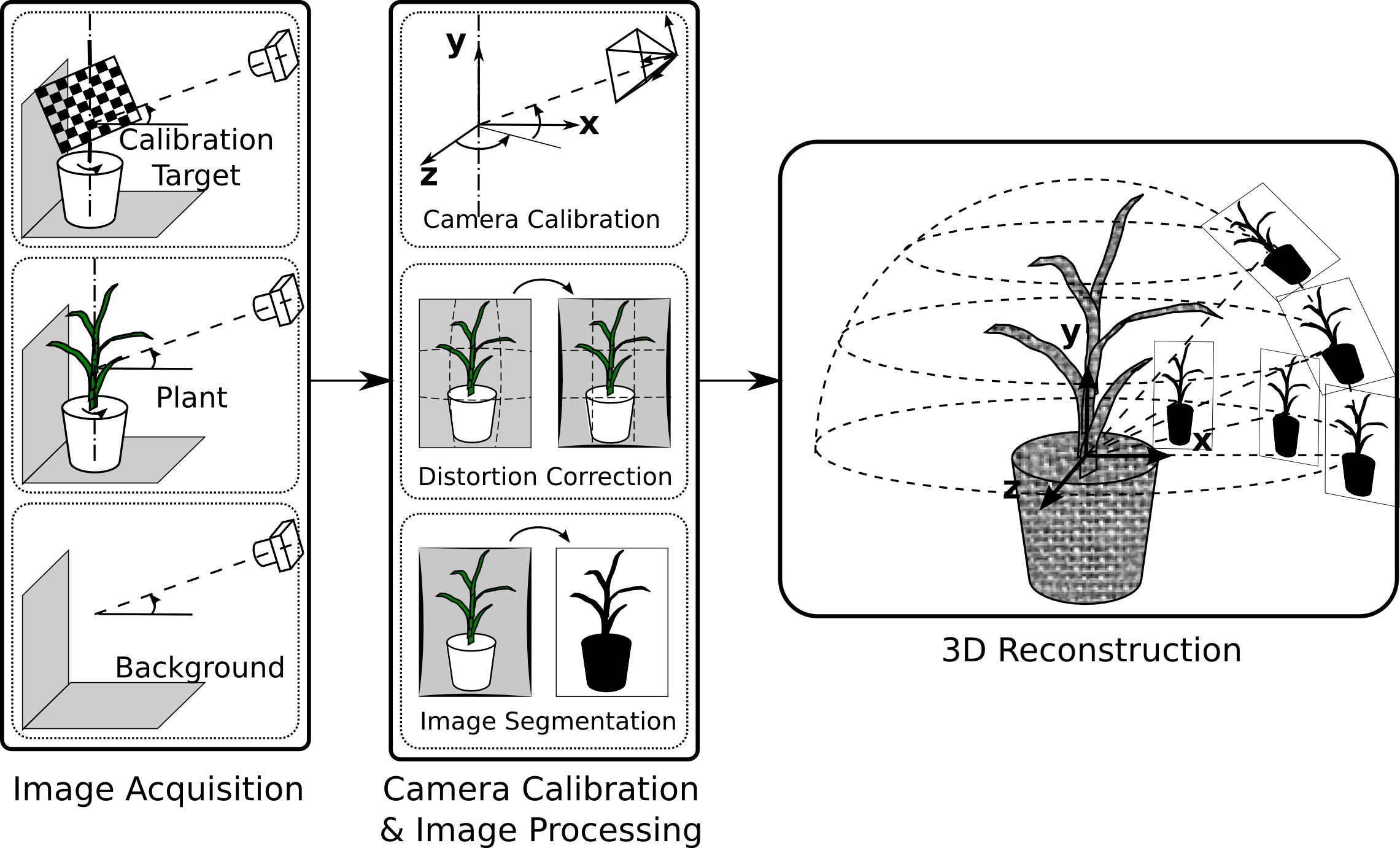}
\caption{Overview of 3D scanning and reconstruction process.}
\label{fig_ProcessOverview}
\end{figure}

PlantScan Lite uses a four step process of 3D scanning and reconstruction (Fig. \ref{fig_ProcessOverview}):
\begin{enumerate}
\item Image acquisition. Multiple-view images of a calibration target, a plant of interest and background are captured using a turntable and tilting cameras.

\item Camera calibration estimates camera parameters for all captured images.

\item Image processing corrects for optical distortion and extract plant silhouettes (background removal). 

\item 3D reconstruction. This paper focuses on the visual hull reconstruction method which uses silhouette images and corresponding camera parameters and poses to create a 3D mesh model. The model is then processed to extract plant geometry information.
\end{enumerate}

\subsection{Image acquisition system}
\begin{figure*}[!t]
\centering
\includegraphics[width=6in]{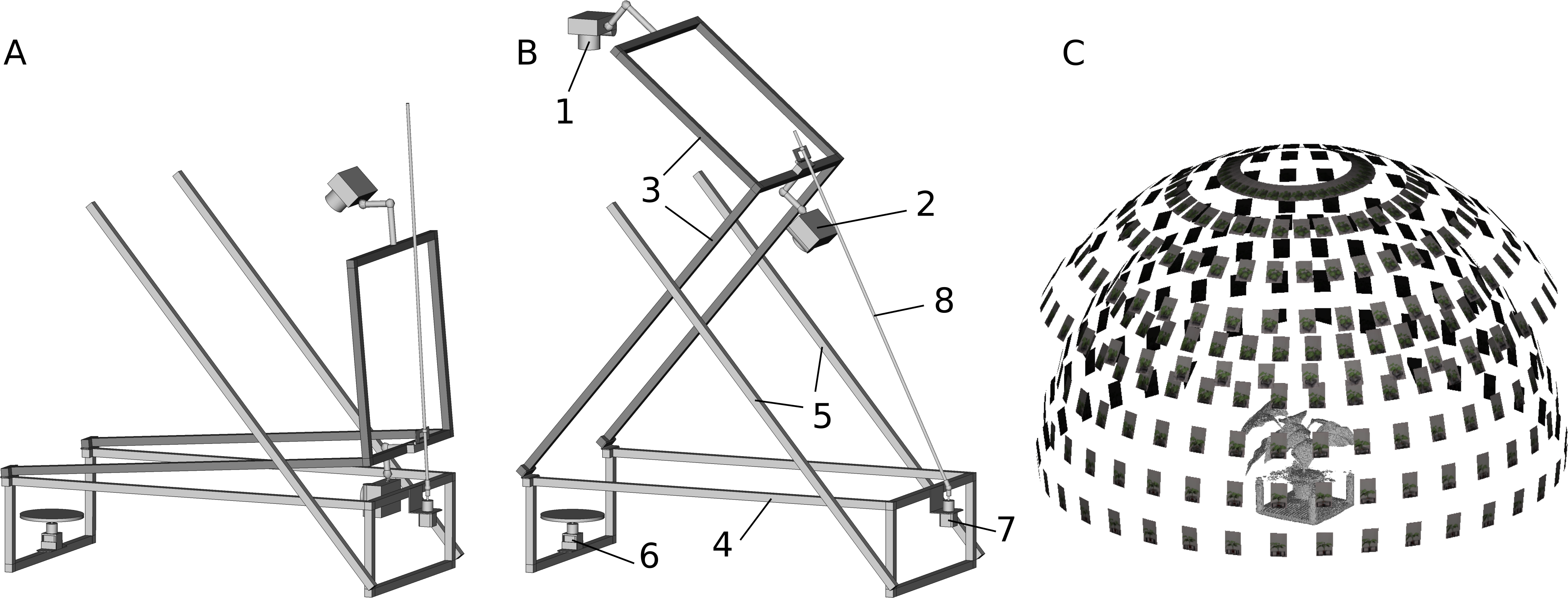}
\caption{Mechanical setup of PlantScan Lite with dual cameras. A) Lowest position - approximately zero degree for the lower camera. B) Highest position - approximately 45 degrees. Two cameras (1, 2) are attached to the tilted frame (3) at approximately 40 degrees apart and jointly cover a full hemisphere scanning surface as in C). A 3D mesh is included to show the size and position of the actual plant relative to the camera view.
}
\label{fig_MiniPlantScanIllustration}
\end{figure*}

PlantScan Lite's image acquisition system consists of (Fig. \ref{fig_MiniPlantScanIllustration}):

\begin{itemize}
\item Two Canon DSLR cameras (1, 2) at an angle of approximately 40 degrees. The cameras are powered by AC adapter kits and connect to a computer through USB cables for tethering capture.

\item Two aluminum frames, one tilted (3) and one fixed to the ground (4). The tilted frame is to mount the cameras and move them up/down. The frames join together with two hinges where the axis of rotation crosses that of the turntable near the middle of a plant to be scanned. Two guiding bars (5) are attached to the lower frame (4) both to keep the tilted frame moving on a vertical plane.

\item Two Phidget bipolar stepper motors (6, 7); one drives a turntable, the other moves the upper frame and the two cameras via a threaded rod (8). Both stepper motors have a gearbox to increase the rotation resolution (0.018 degree/step for the turntable) and torque (18 kg$\cdot$cm for the threaded rod). The turntable controller is synchronised with the cameras via software based on Phidget Python library \cite{Phidgets2016} and Piggyphoto\cite{Piggyphoto2016}.
\end{itemize}

\subsection{Camera calibration for turntable}

\begin{figure}[!t]
\centering
\includegraphics[width=3.25in]{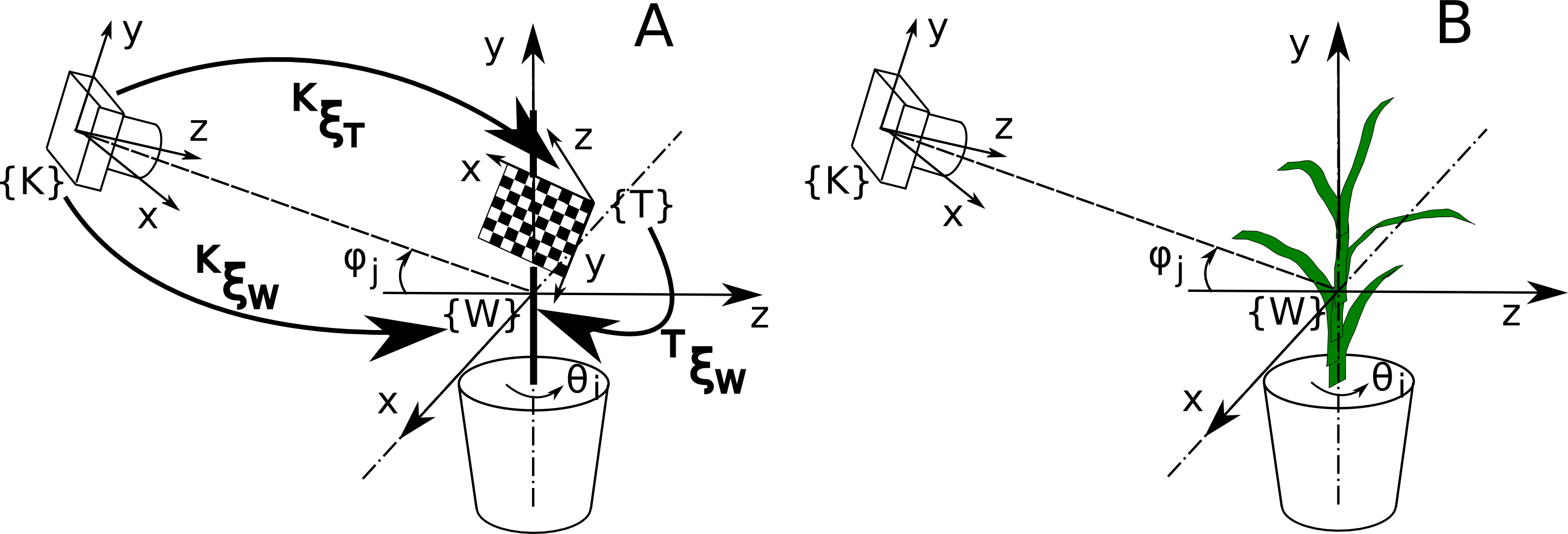}
\caption{Left: A camera K at tilt position j has its own coordinate system $\{K\}$ with a relative pose ${}^K\xi_W$ to the world coordinate system $\{W\}$ attached to the turntable. The camera is panned around relative to turntable at angle $\theta_i$. The target has its own coordinate system $\{T\}$ with a relative pose ${}^T\xi_W$ to the world. Right: a plant is scanned in the same way as the chessboard.
}
\label{fig_CameraCalibration}
\end{figure}

Fig. \ref{fig_CameraCalibration} shows the schematic description of the system with a calibration target and a plant (notation follows \cite{Corke2011}). A chessboard target has a local coordinate system $\{T\}$ with relative pose ${}^T\xi_W$  to the world $\{W\}$ which is fixed to the turntable. A camera at tilt position j has a local coordinate system $\{K\}$ and relative pose ${}^K\xi_W$  to the world coordinates $\{W\}$. Standard mono camera calibration only provides camera pose ${}^K\xi_T$ relative to target at angles where the chessboard pattern is visible to the camera. The aim of camera calibration for the turntable is to find intrinsic parameters and poses ${}^K\xi_W$ for all cameras at all positions. Three constraints are considered: a) turntable rotation angle is known, b) images captured by the same camera have the same intrinsic parameters, and c) different extrinsic parameters are assigned to different tilt positions as if these belong to different cameras that form a vertical arc. The calibration process consists of (1) mono and stereo camera calibrations (2) plane and circle fitting (3) camera to world pose derivation, and (4) optimisation of all camera parameters. Theses step are detailed below.

\subsubsection{Mono and stereo camera calibration}

\begin{figure}[!t]
\centering
\includegraphics[width=3.25in]{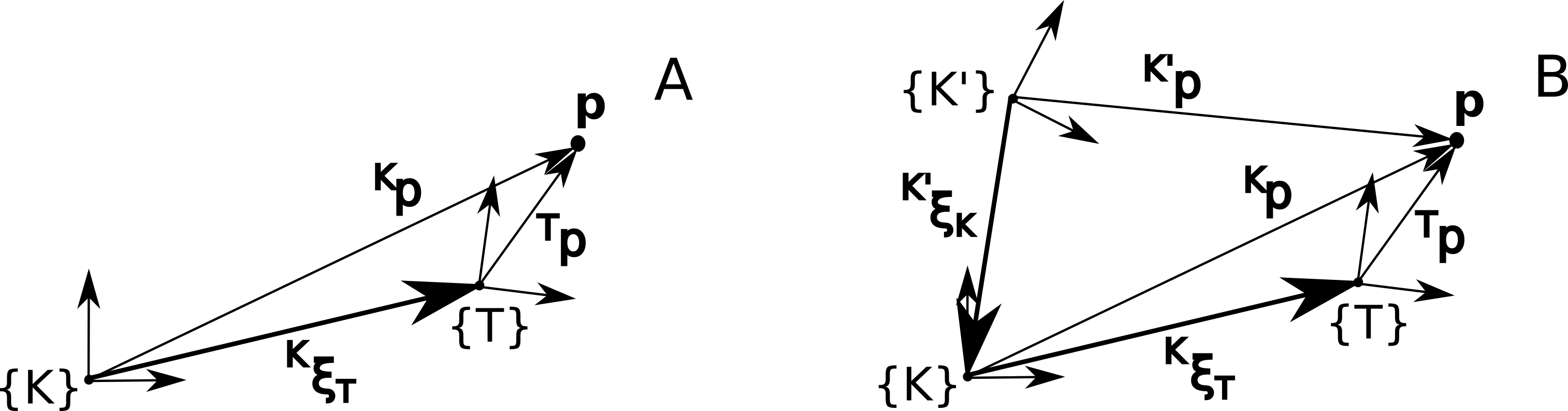}
\caption{Existing camera calibration available in OpenCV library. A) shows camera pose ${}^K\xi_T$ relative to target obtained from mono camera calibration. Point \textbf{p} has relative coordinates ${}^{K}p$ to $\{K\}$, and ${}^{K'}p$ to $\{K'\}$ as in B).
}
\label{fig_CameraCalibrationOpenCV}
\end{figure}

The OpenCV library provides mono and stereo camera calibration routines \cite{Bradski2000, OpenCV2016}. The calibration procedure starts by taking images of a chessboard target at different angles and distances. Corners of the squares on chessboard target are detected. To make this detection efficient, images of the chessboard calibration target are repeatedly scaled down 50\% in a pyramid fashion to approximately 1K $\times$ 1K resolution at the top level. The detected corner positions are obtained at the top level (lowest resolution) using OpenCV's function \textit{findChessboardCorners} and tracked with subpixel accuracy using function \textit{cornerSubPix} on images at lower levels of the pyramid. 

Fig. \ref{fig_CameraCalibrationOpenCV} describes transformations between different coordinate systems. Thin arrows represent coordinate vectors. A pair of orthogonal arrows denotes a coordinate system. Thick arrows show relative pose between two coordinate systems. The coordinates (in mm) of a corner \textbf{p} with respect to the chessboard target coordinate system $\{T\}$ are represented as $^T p = [X, Y, Z=0, 1]^T$. The position of \textbf{p} relative to a camera coordinate system $\{K\}$ is $^K p = [x, y, z]^T$. The relationship between $^T p$ and $^K p$ is expressed as:
\begin{equation}
{}^Kp = {}^K\xi_T \cdot {}^Tp = [R | t] \cdot [X, Y, Z, 1]^T
\end{equation}
where $R$ is a $3 \times 3$ rotation matrix, $t$ a $3 \times 1$ translation vector. This rotation translation matrix $[R | t]$ represents the extrinsic parameters of camera K relative to target T.  The rotation matrix can be represented as a $3 \times 1$ angle-axis rotation vector, so extrinsic parameters have only 6 independent components $[r_0, r_1, r_2, t_0, t_1, t_2]$. The ``$\cdot$'' operator is a matrix-to-vector multiplication.

Point \textbf{p} forms an image on camera sensor at coordinates ${}^Ip = [u, v]^T$. An extended pinhole camera model is used to describe this relationship between ${}^Kp$ and ${}^Ip$ with radial optical distortion:
\begin{equation}
r = \sqrt{\left(\frac{x}{z}\right)^2 + \left(\frac{y}{z}\right)^2}
\label{eq_radius}
\end{equation}
\begin{equation}
u = \frac{f x}{z}(1 + d_1 r^2 + d_2 r^4) + c_u
\label{eq_u}
\end{equation}
\begin{equation}
v = \frac{f y}{z}(1 + d_1 r^2 + d_2 r^4) + c_v
\label{eq_v}
\end{equation}
 where $f$ is focal length, $[c_u, c_v]$ optical center on image, and $[d_1, d_2]$ radial distortion coefficients. A vector $[f, c_y, c_v, d_1, d_2]$ represents intrinsic camera parameters. 

To calibrate the camera, multiple images of the same target are captured at different pan angles $\theta_i$ and tilt angles $\phi_j$, where i = 0 to $i_{max}$ and j = 0 to $j_{max}$. Mono camera calibration, using OpenCV's \textit{calibrateCamera} (based on \cite{Zhang2000}), takes lists of ${}^Tp$ and ${}^Ip$ and computes intrinsic parameters $[f, c_y, c_v, d_1, d_2]_k$ for the camera, and extrinsic parameters $[R_{ijk} | t_{ijk}]$ for each of the images. 

Fig. \ref{fig_CameraCalibrationOpenCV}B shows the multiple camera setup with an additional camera $K'$ at a different tilt position. The same point \textbf{p} is seen by the camera $K'$ at coordinates ${}^{K'}p$. The transformation between the two camera coordinate systems gives: 
\begin{equation}
{}^{K'}p = {}^{K'}\xi_K \cdot {}^Kp = {}^{K'}\xi_K \otimes {}^K\xi_T \cdot {}^Tp
\end{equation}

The transformation ${}^{K'}\xi_K$ between cameras K and $K'$ is equal to stereo transformation $[R_{k,k'} | t_{k,k'}]$ (using OpenCV's \textit{stereoCalibCameras}). If K and $K'$ are of the same camera but at different tilting angles $\phi_j$ and $\phi_{j+1}$, the transformation becomes $[R_{j,j+1} | t_{j,j+1}]$. The ``$\otimes$'' operator denotes matrix-to-matrix multiplication. The transformation can be applied repeatedly between successive camera pairs at different tilt positions. Given stereo transformations between successive cameras and the pose of the first camera, the poses of other cameras are also obtained.

\subsubsection{Estimation of axis and centre of rotation}

\begin{figure}[!t]
\centering
\includegraphics[width=3.25in]{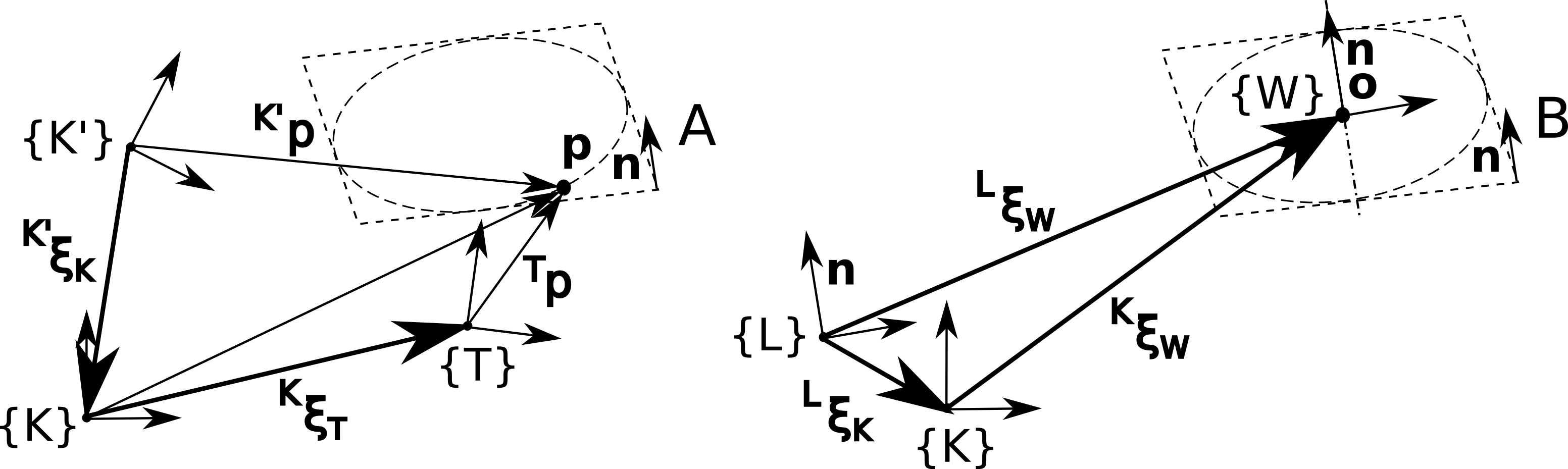}
\caption{Rotation axis estimation from target movement. A) A point \textbf{p} on calibration target moves on a planar circular orbit with a normal vector n as the target rotates. B) Use a new coordinate $\{L\}$ that is parallel to normal vector n to locate the world coordinate at the centre of orbit. Relative pose  is computed.
}
\label{fig_RotAxisEstimation}
\end{figure}

Location of the world coordinate system fixed to the rotation axis is found in two steps (Fig. \ref{fig_RotAxisEstimation}): (1) compute the normal vector \textbf{n} of the plane containing the rotation orbit of the target's corners, and (2) fit a circle to find the centre of rotation \textbf{o}. The normal vector and centre of rotation define the world coordinates.
%
%
%

First, rotation axis is estimated from the rotational motion of the chessboard target. From extrinsic parameters $[R_{ijk} | t_{ijk}]$, the same chessboard corner ${}^Tp$ is seen as ${}^Kp = [R_{ijk} | t_{ijk}] {}^Tp$ moving on a circular orbit as shown in Fig. \ref{fig_RotAxisEstimation}A. Note that the chessboard pattern and its positions can only be detected in 1/4 to 1/3 of the number of target images around the circular orbit. The equation of the orbit plane is expressed as:
\begin{equation}
a(t_x - x_m) + b(t_y - y_m) + c(t_z - z_m) = 0
\end{equation}
where $[x_m, y_m, x_m]$ is the centroid (or the mean) of all positions ${}^Kp$ on the same orbit, and $[a, b, c]$ is the normal vector of the plane.

Equation (6) can be rewritten in matrix form $\mathbf{Bn} = 0$ where $\mathbf{n} = [a, b, c]$ and $\mathbf{B}$ is a matrix containing chessboard positions relative to the centroid of all the positions. Vector $\mathbf{n}$ is an eigenvector corresponding to the smallest eigenvalue obtained from Singular Value Decomposition of matrix $\mathbf{B}$. 

Second, to find the centre of rotation of the calibration target, a different coordinate system $\{L\}$ is used (Fig. \ref{fig_RotAxisEstimation}B). $\{L\}$ is in fact equivalent to $\{K\}$ with a rotation transformation ${}^L \xi_K$ such that the y-axis is parallel to $\mathbf{n}$. In $\{L\}$, a 2D circle can be fitted onto the target point orbit and the centre of rotation can be obtained. The relationship between $\{L\}$ and $\{K\}$ with respect to $\{W\}$ is:
\begin{equation}
{}^L\xi_W = {}^L\xi_K \otimes {}^K\xi_W
\end{equation}

The transformation ${}^L\xi_W$ has the form $[R_{\omega} | 0]$, where $R_{\omega}$ is a rotation matrix whose angle-axis rotation vector $\omega \mathbf{w} = [\omega w_x, \omega w_y, \omega w_z]^T$ can be obtained as:

\begin{equation}
\omega = \arctan \left( \frac{|\mathbf{n} \times \mathbf{y}|}{\mathbf{n} \cdot \mathbf{y}} \right)
\end{equation}

\begin{equation}
\mathbf{w} = \frac{\mathbf{n} \times \mathbf{y}}{|\mathbf{n} \times \mathbf{y}|}
\end{equation}
where $\omega$ is the rotation angle and $\mathbf{w}$ is a vector around which the rotation is applied to turn $\mathbf{n}$ to y-axis $\mathbf{y}$ of $\{W\}$. The bar denotes vector normalization.

$R_{\omega}$ is computed from $\omega \mathbf{w}$ by Rodrigues' formula:
\begin{equation}
R_{\omega} = \cos \omega I + (1 - \cos \omega)\mathbf{w} \mathbf{w}^T + \sin \omega [\mathbf{w}]_{\times}
\label{eq_Rodrigues}
\end{equation}

After applying the rotation transformation $R_{\omega}$ to target positions, a circle can be fitted to $[z, x]^T$ coordinates by a Linear Least-Squares algorithm \cite{Coope1993}. This fitting gives the centre of the orbit $[x_0, y_0, z_0]^T$, with $y_0$ to be the averaged y-component of the target point positions in $\{L\}$ coordinate system.

Now the world coordinate system is set at centre of rotation $\mathbf{o}$ and with its axes parallel to those of $\{L\}$, the transformation from $\{L\}$ to $\{W\}$ is:
\begin{equation}
{}^L\xi_W = [I | t_0]
\end{equation}
where $t_0 = [x_0, y_0, z_0]^T$. As a result, pose of camera K relative to $\{W\}$ can be expressed as:

\begin{eqnarray}
{}^K\xi_W & = & {}^K\xi_L \otimes {}^L\xi_W = \left( {}^L\xi_K \right)^{-1} \otimes {}^L\xi_W \\
& = & [R_{\omega} | 0]^{-1} [I | t_0]
= [R^T_{\omega} | R^T_{\omega} t_0]
\end{eqnarray}

\subsubsection{Estimation of camera poses relative to the world coordinate system fixed to the turntable axis}

\begin{figure}[!t]
\centering
\includegraphics[width=3.25in]{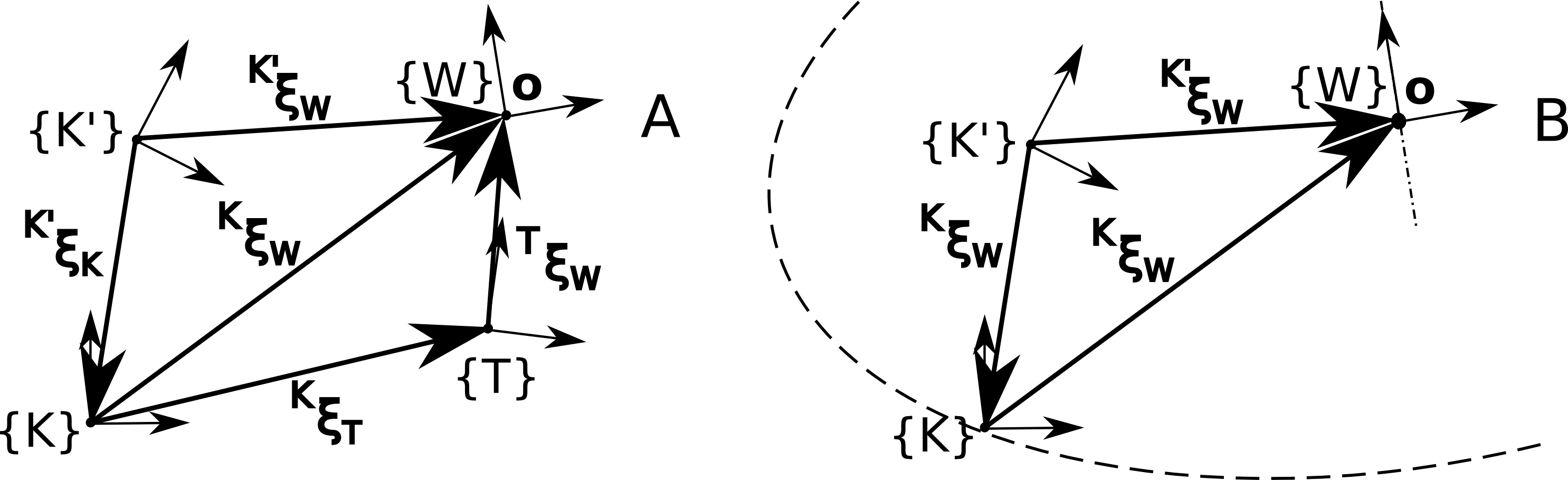}
\caption{Camera pose with respect to world coordinate system. A) Relative poses of ${}^K\xi_W$ and ${}^{K'}\xi_W$ are obtained. B) Camera coordinate $\{K\}$ moves on a circle centre at $\{W\}$ origin.
}
\label{fig_PoseEstimation}
\end{figure}

Since the pose ${}^K\xi_W$ of the camera $K$ at tilt position $\phi_j$ is obtained, the pose of any additional camera $K'$ relative to the world coordinate system can be obtained from a given stereo transformation ${}^K\xi_K$  as shown in Fig. \ref{fig_PoseEstimation}A:
\begin{equation}
{}^{K'}\xi_T = {}^{K'}\xi_K \otimes {}^K\xi_W
\end{equation}
Similarly, the pose of the target relative to the world coordinate system is expressed as:
\begin{equation}
{}^T\xi_W = {}^T\xi_K \otimes {}^K\xi_W = \left( {}^K\xi_T \right)^{-1} \otimes {}^K\xi_W
\end{equation}

We are interested in the \textit{reverse} transformation to obtain target point position in the world coordinate system:
\begin{eqnarray}
{}^W\xi_T & = & {}^W\xi_K \otimes {}^K\xi_T = \left( {}^K\xi_W \right)^{-1} \otimes {}^K\xi_T \\
& = & [R_{ijk} | t_{ijk}]^{-1} [R^T_{\omega} | R^T_{\omega} t_0] \\
& = & [R^T_{ijk} R^T_{\omega} | R^T_{ijk} R^T_{\omega} t_0 - R^T_{ijk} t_{ijk}]
\end{eqnarray}

Since there are multiple estimates of ${}^W\xi_T$ for different rotation angle $\theta_i$, ${}^W\xi_{T0}$ for zero rotation angle is obtained by applying an inverse of the rotation to the corresponding pose:
\begin{eqnarray}
{}^W\xi_{T0} & = & [R^{-1}_{\theta_i} | 0] {}^W\xi_T \\
& = & [R^T_{\theta_i} R^T_{ijk} R^T_{\omega} | R^T_{\theta_i} (R^T_{ijk} R^T_{\omega} t_0 - R^T_{ijk} t_{ijk})]
\end{eqnarray}
where $R_{\omega}$ is the matrix obtained from angle-axis rotation vector (equation (\ref{eq_Rodrigues})), and $R^{-1}_{\theta_i} = R^T_{\theta_i}$.

Since the world coordinate systems and the target are fixed, the camera coordinate system needs to move in a circle around y axis of $\{W\}$ to represent the correct relative motion seen by the camera, as shown in Fig. \ref{fig_PoseEstimation}B. The pose of camera K for rotation angle $\theta_i$ is:
\begin{eqnarray}
{}^{K\theta_i}\xi_W & = & {}^{K}\xi_W [R_{\theta_i} | 0] = [R^T_{\omega} R_{\theta_i} | R^T_{\omega} t_0]
\end{eqnarray}

The pose of camera $K'$ for rotation angle  is :
\begin{eqnarray}
{}^{K'\theta_i}\xi_W & = & {}^{K'\theta_i}\xi_W [R_{\theta_i} | 0] \\
& = & [R_{j,j+1} R^T_{\omega} R_{\theta_i} | R_{j,j+1} R^T_{\omega} t_0 + t_{j,j+1}]
\end{eqnarray}
 
\subsubsection{Optimisation to refine camera parameters}
A nonlinear least-square optimisation is applied to refine estimates of camera intrinsic parameters $[f, c_u, c_v, d_1, d_2]$ and angle-rotation vector and translation vector of camera pose ${}^K\xi_W$ and ${}^{K'}\xi_W$ at different tilt angles, and target inverse pose ${}^W\xi_{T0}$. Optimisation seeks to minimize is the pixel distance between projected target corners to camera and the corners on the actual images. 

Estimated position of chessboard corner relative to camera K at rotation $\theta_i$: 
\begin{equation}
{}^Kp_{estimated} = {}^{K\theta_i}\xi_W \otimes {}^W\xi_{T0} \cdot {}^Tp
\end{equation}
A pinhole camera projection with radial distortion is applied with equations (\ref{eq_radius}), (\ref{eq_u}) and (\ref{eq_v}) to obtain corresponding image point ${}^{I\theta_i}p_{estimated}$. This is applied to other tilt positions and the second camera. The squared distance between image projection  ${}^{I\theta_i}p_{estimated}$ of estimated corners and their detected image positions is minimised.

\subsection{Background subtraction}
Plants with large leaves can cast strong shadows, so a simple image threshold will not completely remove the background. We found that a shadow removal algorithm based on static background proposed in \cite{Lo2006} performs  background removal  for thin leaves more accurately and with less computation than other techniques \cite{KaewTraKulPong2002, Zivkovic2004, Zivkovic2006}. Here, we extend the technique of \cite{Lo2006} for LAB color space \cite{Wikipedi2016}, further improving  background removal accuracy. 

Suppose $L(u,v)$, $A(u,v)$ and $B(u,v)$ are the luminance and two color channels of the current image and $L'(u,v)$, $A'(u,v)$ and $B'(u,v)$ the corresponding image channels of the background image. Three error functions applied to each pixel position $[u,v]$ are defined as follows:
\begin{equation}
\Delta(u,v) = \left| L(u,v) - L'(u,v) \right|
\end{equation}
\begin{multline}
\Theta(u,v) = \left| A(u,v) - A'(u,v) \right| + \\
\left| B(u,v) - B'(u,v) \right|
\end{multline}
\begin{multline}
\Psi(u,v) = \left| \frac{L(u,v)}{L(u+1,v)} - \frac{L'(u,v)}{L'(u+1,v)} \right| + \\
\left| \frac{L(u,v)}{L(u,v+1)} - \frac{L'(u,v)}{L'(u,v+1)} \right|
\end{multline}
These error functions $\Delta(u,v)$, $\Theta(u,v)$ and $\Psi(u,v)$ represent the differences in luminance, color and texture respectively. An overall score is computed to determine a pixel as foreground or background:

\begin{equation}
\Omega(u,v) = \frac{\alpha \Delta(u,v) + \beta \Theta(u,v) + \gamma \Psi(u,v)}{\alpha + \beta + \gamma}
\end{equation}
where the values of $\alpha$, $\beta$, $\gamma$ are found empirically. 

\begin{figure}[!t]
\centering
\includegraphics[width=3in]{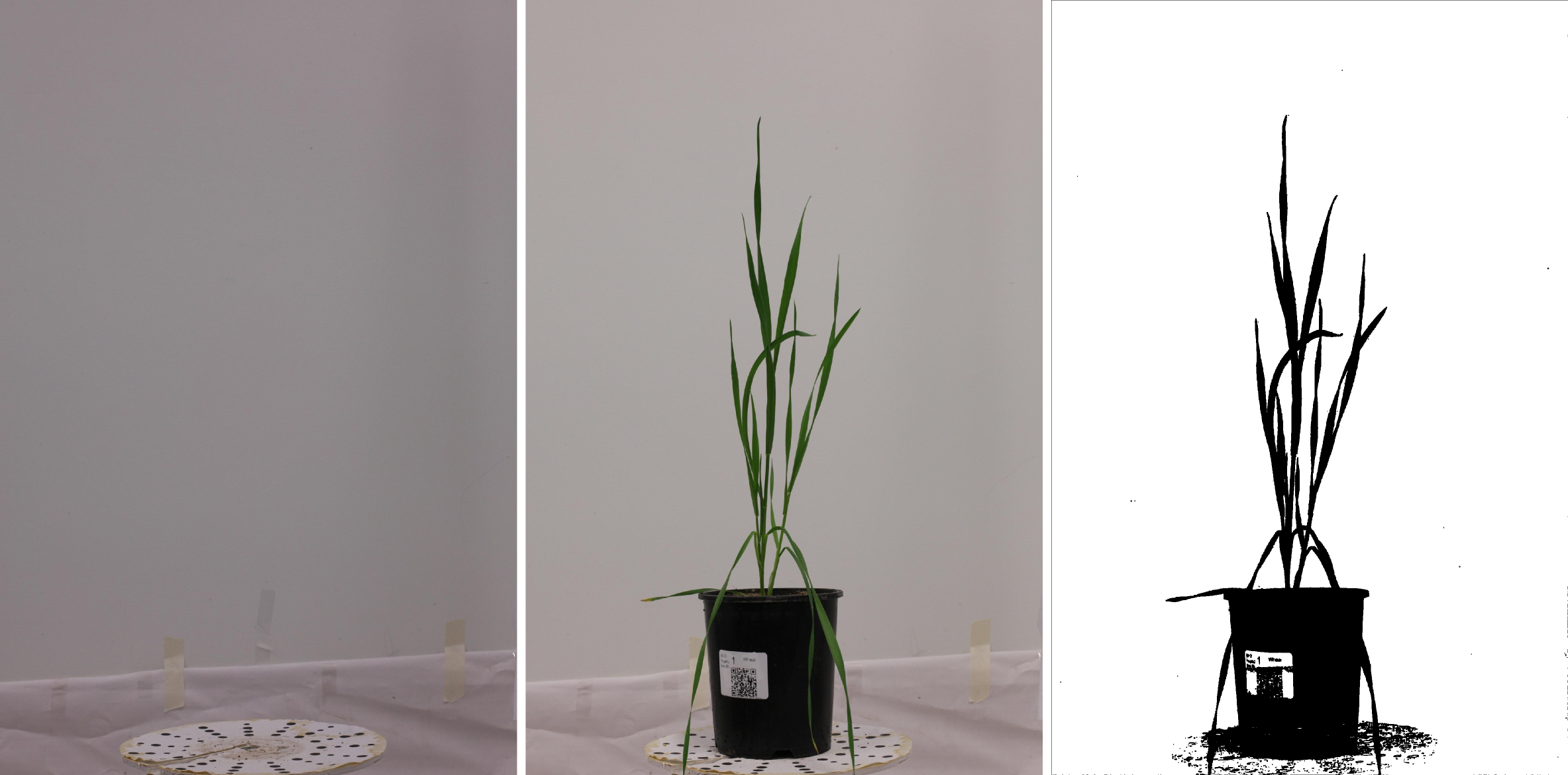}
\caption{Removal of background (left) from a plant image (middle) to produce a silhouette image (right).}
\label{fig_ImageSegmentation}
\end{figure}

A threshold $t$ is applied to $\Omega(u,v)$ to separate background and foreground. For our images, $\alpha = 0.1$, $\beta = 0.5$, $\gamma = 0.4$ and $t = 5$ to 10 were found to work well. Unlike \cite{Lo2006}, our proposed technique allows for segmentation of dark objects (such as the plant pots) and this is controlled via coefficient $\alpha$. Fig. \ref{fig_ImageSegmentation} shows an example of background removal using our proposed algorithm.

\subsection{3D reconstruction}

\subsubsection{Bounding box estimation}

\begin{figure}[!t]
\centering
\includegraphics[width=1.5in]{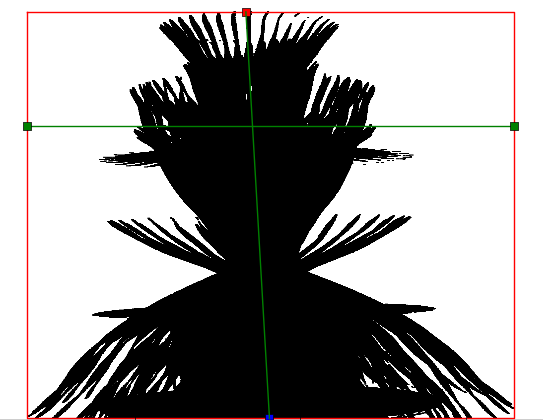}
\caption{Initial bounding box estimate from the overlapping of silhouettes from horizontal camera. The vertical green centreline represents the projected turntable axis. }
\label{fig_BoundingBox}
\end{figure}

The bounding box of the subject is obtained in two steps:

\begin{enumerate}
\item An initial 3D bounding box is estimated based on silhouettes from the most horizontal camera view. These silhouette images are overlapped/combined into a single image and 2D bounding box is computed (Fig. \ref{fig_BoundingBox}). The Y axis and the origin are projected onto this overlapped image. The crossing points of the projected axis with the 2D bounding box are mapped back to the turntable axis in 3D space to obtain $y_{min}$ and $y_{max}$. The back projection of the rectangle width to the world origin gives a single magnitude for $x_{min}$, $x_{max}$, $z_{min}$ and $z_{max}$.

\item A refined bounding box is calculated from a 3D reconstruction at a low resolution ($128^3$ voxels) using the initial bounding box. This takes only a few seconds to compute. Particularly, we found that no thin parts of the plant are missing when reconstructed at low resolution. As a result, the refined bounding box tightly contains the 3D space of the plant.
\end{enumerate}

\subsubsection{Volume reconstruction}

\begin{figure}[!t]
\centering
\includegraphics[width=3in]{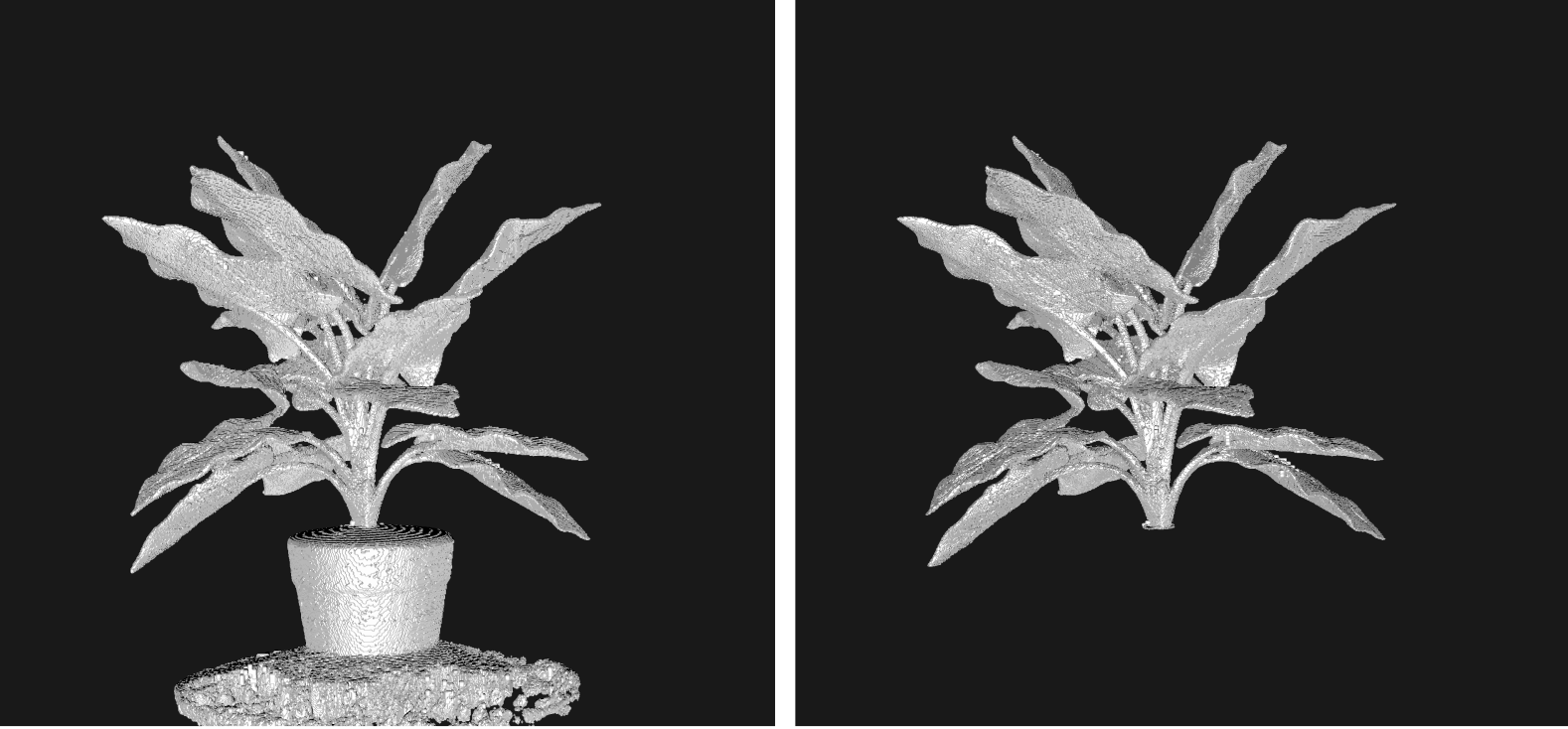}
\caption{3D mesh reconstruction without and with pot and pot carrier subtraction.}
\label{fig_PotRemoval}
\end{figure}

In this work, the 3D plant was reconstructed using a visual-hull volume carving approach. This method recovers sharp and thin structures common to plants (although a major drawback is that it cannot correctly recover concave surfaces, making reconstructions of curved surfaces such as leaves thicker than they should be). There may be plant movements induced by air circulation or mechanical vibration which can be accounted for by some tuning during reconstruction. The reconstruction method consisted of 3 steps:
\begin{enumerate}
\item A 3D volume equal to the bounding box is generated and split into voxels. Each voxel is repeatedly projected into the silhouettes and its 2D signed distance to the nearest boundary of each silhouette is calculated. If the distance is negative (outside) in any of the silhouettes, the voxel is flagged as empty. To accommodate some uncertainty in the silhouettes and plant movements, a voxel is set to be removed if it is outside more than a fixed number of silhouettes (3 is chosen in this paper). The process repeats until the end where remaining voxels form a 3D hull model of plant. An octree structure is used for voxel removal from lowest resolution to the highest resolution \cite{Szeliski1993}, giving a 3X speedup as compared to processing all voxels of a full resolution 3D volume.

\item Removal of pot and pot carrier. Since we are only interested in the plant, the pot and pot carrier need to be removed to simplify mesh analysis as well as reduce the mesh size. One method is to subtract voxels inside a given bounding tapered cylinder of the pot and pot carrier. Fig. \ref{fig_PotRemoval} shows the snapshot of the mesh of one of the plastic plants before and after pot subtraction. This needs to be done before the 3D meshing step to produce a clean and watertight mesh.

\item 3D meshing by marching cube from the remaining voxels. A grid point is checked against 8 surrounding voxels. The value of the 8 surrounding voxels is matched with a 256-element lookup table to determine if the grid is on or close to the mesh so that a polygon can be created from this grid and nearby grid points.
\end{enumerate}

\subsection{Mesh segmentation and geometry}

\begin{figure}[!t]
\centering
\includegraphics[width=3in]{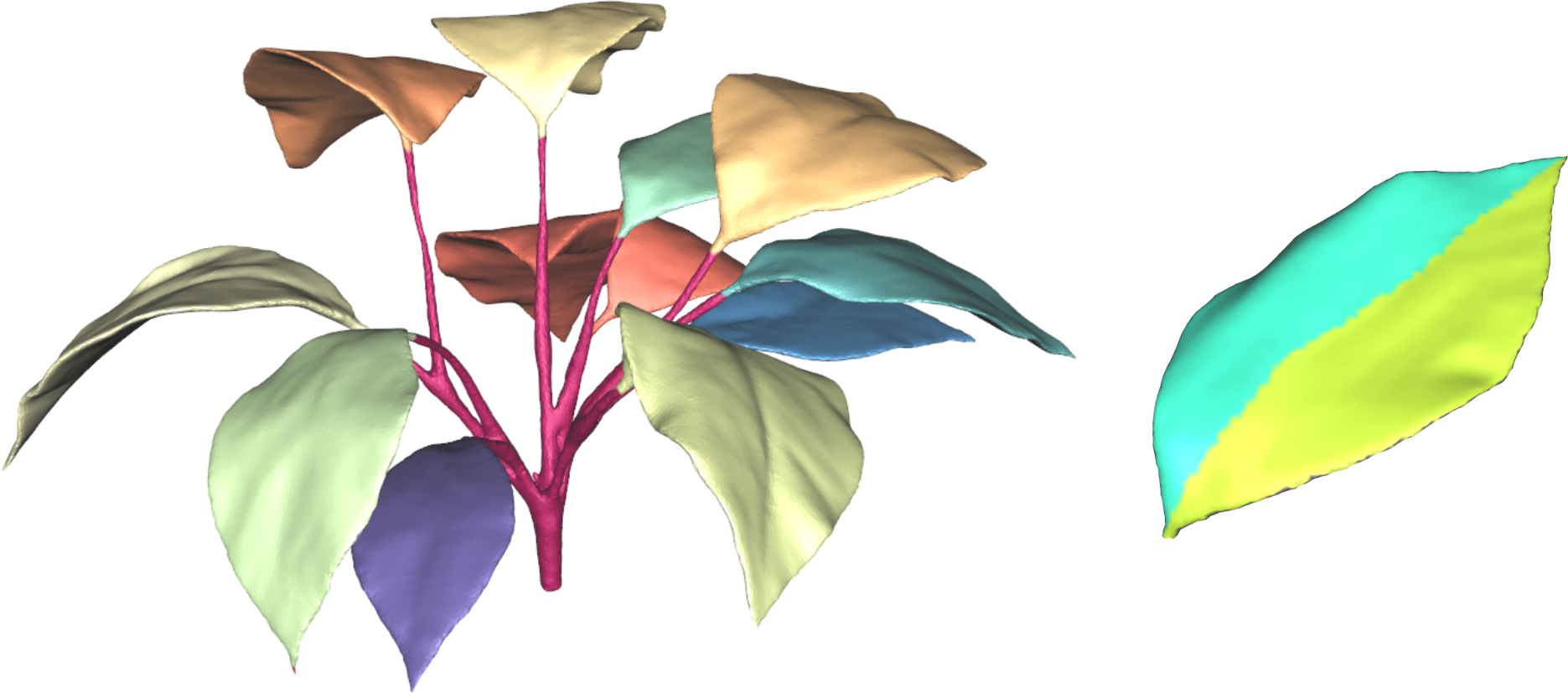}
\caption{Left: snapshot of 3D leaf segmentation of the big plastic plant to measure their geometry. Right: snapshot of the centreline extracted from a single leaf to estimate the length. The width is calculated based on the distance between the two extremities across this axis.}
\label{fig_MeshSegmentation}
\end{figure}

Mesh segmentation algorithms involve assigning a unique label to all the vertices of the 3D mesh that belong to the same region. This paper uses a simplified version of the ``hybrid'' segmentation pipeline previously presented in \cite{Paproki2012}. Primarily it is based around a constrained region-growing algorithm. In short, the curvature and normal for the 3D mesh were pre-computed. A user defined curvature threshold was provided to find large ``flat'' regions (e.g., broad leaves) for use as seed regions. A curvature-constrained region growing was then performed from each seed region. The geometry (area, width, length, and circumference) of each segmented leaf was then extracted using the approach outlined in \cite{Paproki2012}. The result for large plastic plant is shown in Fig. \ref{fig_MeshSegmentation}. 

\section{Results}
\begin{figure}[!t]
\centering
\includegraphics[width=3.5in]{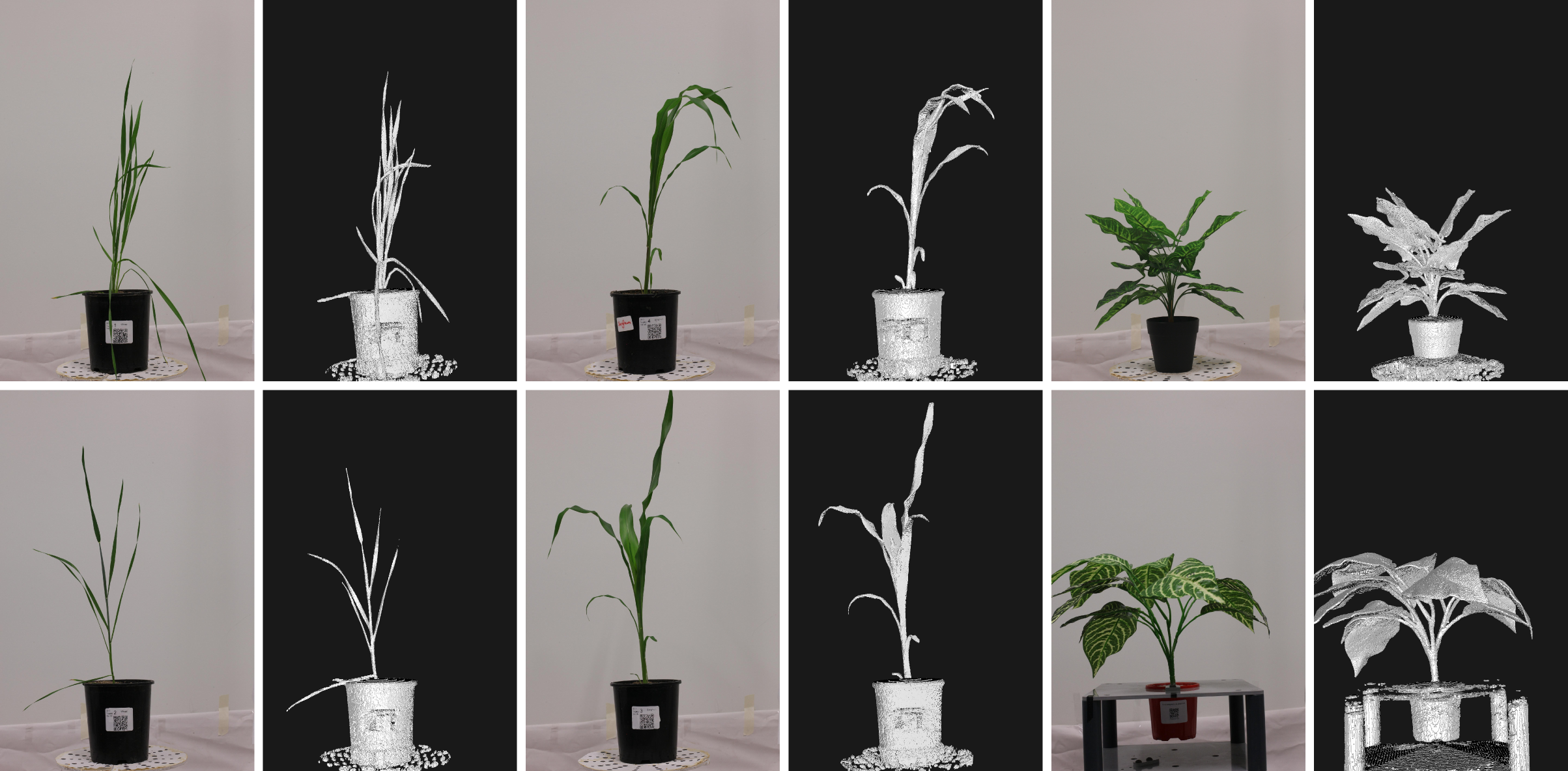}
\caption{Examples of 3D reconstruction of Wheat (Triticum aestivum) (left columns) and Sorghum (Sorghum bicolor) (middle columns) at 5 weeks after sowing and plastic plants (right columns) using 360 images/plant. Color images are one of input images corrected for lens distortion. Grayscale images are rendering of the corresponding 3D models from the same view.
}
\label{fig_ReconstructionSnapshot}
\end{figure}

Fig. \ref{fig_ReconstructionSnapshot} shows reconstructions of different plants with different complexity and leaf shapes. 3D meshes of plants with thin and narrow leaves are reconstructed with excellent geometric agreement, although there is a minor discrepancy at the tips of the leaves. For visual comparison the pots are included in this figure, however they are removed (as in Fig. \ref{fig_PotRemoval}) before geometry measurement. 

The same plastic plants are also reconstructed twice with different numbers of input images to see how this affects the reconstruction quality. Without tilting the cameras, it took 3 minutes for two cameras to capture a total of 72 images, as compared to 30 minutes to capture 360 images where the two camera were moved to 5 tilt positions (taking half of the total scanning time). A visual comparison (not shown in paper) does not show obvious differences between the two meshes of the same plant. 

The leaves of the large plastic plant (bottom of Fig. \ref{fig_PotRemoval}) were dissected and scanned to measure the length, width, perimeter and area). There are 12 leaves grouped into 3 sizes shown in Fig. \ref{fig_LeafSizes} and Tab. \ref{table_LeafSizes}. Since the leaves mostly curve along their length, the length measurement is likely to be affected. For validation of the reconstruction accuracy, the width of the leaves is chosen as this is less affected by the curving. 

\begin{figure*}[!t]
\centering
\subfloat{
\includegraphics[width=2in]{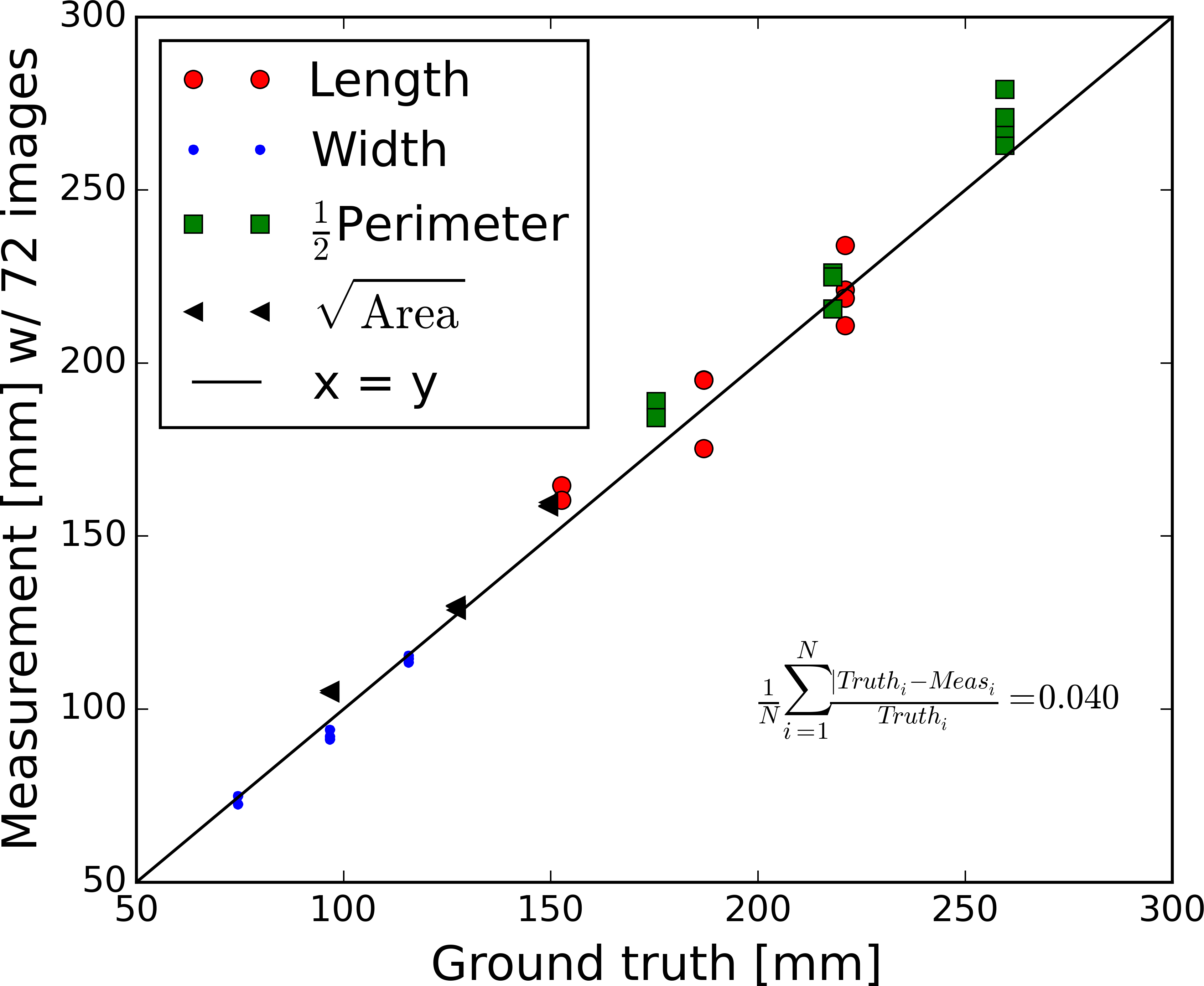}%
\label{fig_first_case}}
\hfil
\subfloat{
\includegraphics[width=2in]{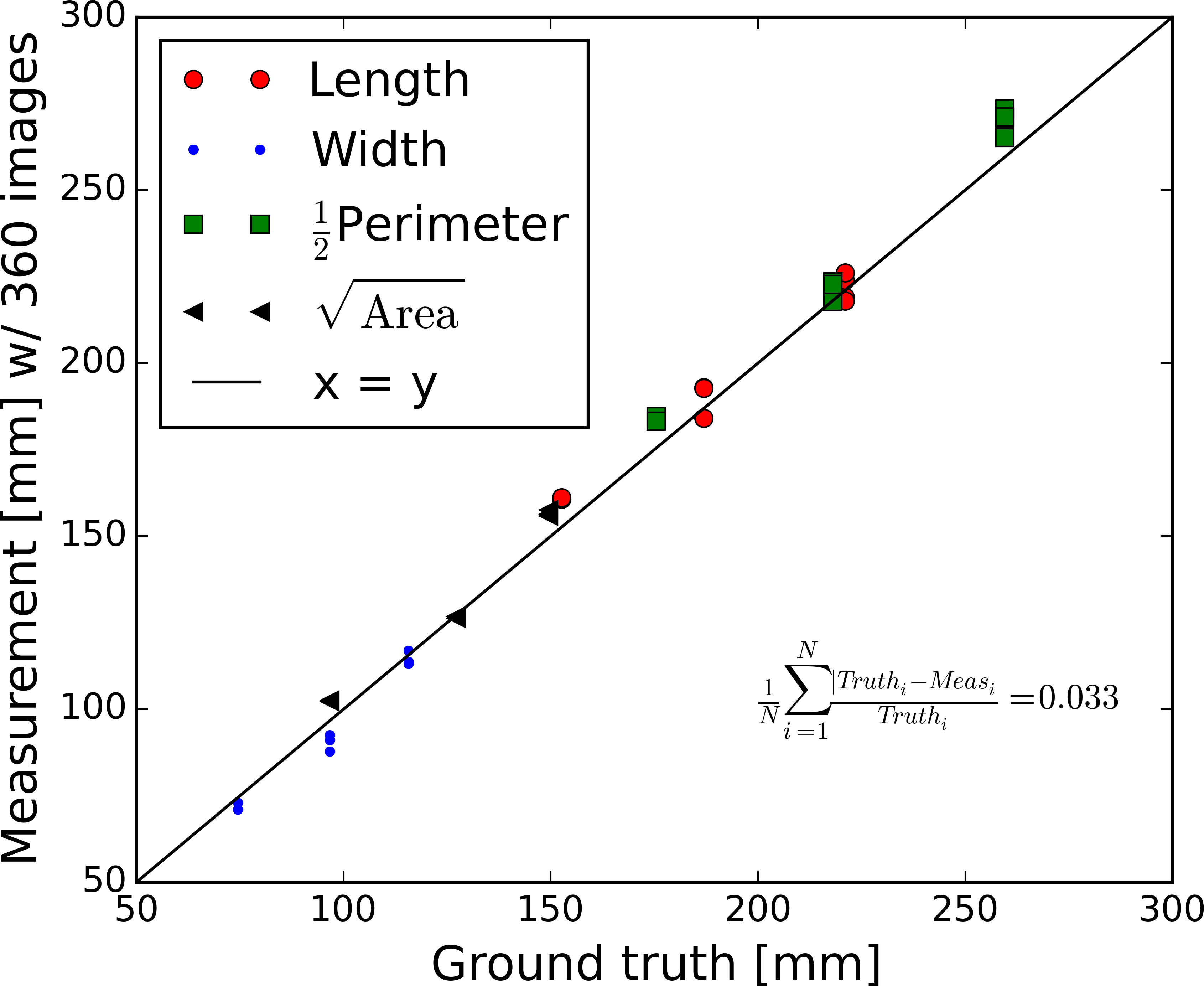}%
\label{fig_second_case}}
\caption{Comparison between ground truth and automatic measurements on 3D models reconstructed from 72 images (left) and from 360 images (right).}
\label{fig_meas_comparison}
\end{figure*}

A quantitative comparison between the two cases is shown in Fig. \ref{fig_meas_comparison}A and B. The ground truth obtained from the 2D scans of the dissected leaves (Fig. \ref{fig_LeafSizes}) was graphed against  the measurement obtained from the 3D meshes of the plant. To fit into the plots, the values of perimeter are scaled down half and the values of area are root-squared. It can be seen that the measurements agree quite well with the ground truth. The average relative error $\epsilon = \frac{1}{N} \sum\limits_{i=0}^{N-1} \frac{|truth_i - measurement_i|}{truth_i}$ of the measurement is 4.0\% for 72 images and 3.3\% for 360 images:

\begin{figure}[!t]
\centering
\includegraphics[width=2in]{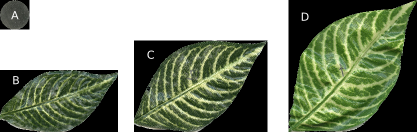}
\caption{Leaf sizes of the large plastic plants. The plastic plant has 4 leaves B, 2 leaves C and 6 leaves D. Circle A has an area of 10 cm$^2$ and diameter of 35.7mm to be used as  reference.}
\label{fig_LeafSizes}
\end{figure}

\begin{table}[!t]
\renewcommand{\arraystretch}{1.3}
\caption{Manual leaf measurements of large plastic plant.}
\label{table_LeafSizes}
\centering
\begin{tabular}{|c||c|c|c|c|}
\hline
Leaf & Length (mm) & Width (mm) & Perim. (mm) & Area 
(mm$^2$)\\
\hline
B & 152.6 & 74.5 & 350.89 & 8535.6\\
\hline
C & 186.9 & 96.6 & 436.16 & 13303.4 \\
\hline
D & 221.0 & 115.6 & 519.1 & 18868.0 \\
\hline
\end{tabular}
\end{table}

\section{Conclusion and discussion}
We have presented a complete system for automatic high-resolution 3D plant phenotyping. Several technical solutions in camera calibration, image processing and 3D reconstruction have been proposed for high accuracy of 3D mesh models. Notably, we proposed a camera calibration procedure that uses a standard chessboard calibration target that is easy to make and use in production environment. We also proposed an extension of foreground segmentation to LAB color space for improved segmentation accuracy for plants with thin leaves commonly found in major crop plants.

The system captures high quality images with accurate camera poses for image-based 3D reconstruction algorithm. The quantitative measurements using 3D visual hull algorithm provided an estimate of the accuracy of the whole system in general. We showed that useful metrics such as leaf width, length and area can be obtained with high accuracy from the 3D mesh models. Fast scanning only takes 3 minutes (72 images) per plant and still produces a reasonable measurement (4\% error). More images (360 images) per plant is required for better accuracy (3.3\% error) especially for complex plant structure, but requires 5 to 10 times more time to scan. 

Future works include a calibration using both pan and tilt axes so that camera pose can be obtained for an arbitrary pair of pan-tilt rotation angles. This would enable a more flexible scanning trajectory other than circular rotation with fixed number of images per tilt angle.

\ifCLASSOPTIONcompsoc
  \section*{Acknowledgments}
\else
  \section*{Acknowledgments}
\fi

Chuong Nguyen acknowledges the support by ARC DP120103896 and CE140100016 through ARC Centre of Excellence for Robotics Vision (http://www.roboticvision.org), CSIRO OCE Postdoctoral Scheme and the National Collaborative Infrastructure Strategy (NCRIS) project, "Australian Plant Phenomics Centre". Thanks to Dr. Geoff Bull of the High Resolution Plant Phenomics Centre for his valuable feedback to the manuscript.



%

\bibliographystyle{IEEEtran}
\bibliography{IEEEabrv,Plantscan-Lite}

\end{document}